\documentclass[10pt,twocolumn,letterpaper]{article}

\usepackage{cvpr}              

\usepackage[dvipsnames]{xcolor}
\usepackage{color, colortbl}
\usepackage{times}
\usepackage{epsfig}
\usepackage{graphicx}
\usepackage{amsmath}
\usepackage{amssymb}

\usepackage{bm}
\usepackage{booktabs}
\usepackage{amsfonts}       
\usepackage{array}
\usepackage{multirow}
\usepackage{xspace}
\usepackage{mathtools}
\usepackage{caption}

\usepackage{dsfont}
\usepackage{multirow}
\usepackage{stmaryrd}
\usepackage{booktabs}
\usepackage{nicefrac}
\usepackage{relsize}
\usepackage{mathabx}
\usepackage{subcaption}
\usepackage{xspace}
\usepackage{array}

\usepackage{mathtools}
\usepackage{tikz}
\usepackage{array}
\newcolumntype{H}{>{\setbox0=\hbox\bgroup}c<{\egroup}@{}}
\usepackage{xspace}
\usepackage{pifont}
\usepackage{tabularx}
\usepackage{amssymb}
\usepackage{hyperref}

\hypersetup{
    colorlinks=true,
    linkcolor=blue,
    }

\definecolor{cvprblue}{rgb}{0.21,0.49,0.74}

\usepackage{tikz}
\newcommand{\xxmark}{%
\tikz[scale=0.23] {
    \draw[line width=0.7,line cap=round] (0,0) to [bend left=6] (1,1);
    \draw[line width=0.7,line cap=round] (0.2,0.95) to [bend right=3] (0.8,0.05);
}}
\def\checkmark{\tikz\fill[scale=0.4](0,.35) -- (.25,0) -- (1,.7) -- (.25,.15) -- cycle;}

\newcommand{\mCross}[0]{\xxmark}

\def\true{1}
\def\showcomments{1} 
\ifx\showcomments\true
  \newcommand{\comments}[1]{\textcolor{blue}{#1}}
\else
   \newcommand{\comments}[1]{}
\fi
\newcommand{\ours}{$\mathrm{Purposer}$\xspace}

\DeclarePairedDelimiter\floor{\lfloor}{\rfloor}

\begin{document}
\newcommand{\GA}[1]{{\color{violet}#1}}
\newcommand{\GPM}[1]{{\color{blue} GPM:#1}} 

\newcommand{\mA}{\mathcal{A}}
\newcommand{\mF}{\mathcal{F}}
\newcommand{\mI}{\mathcal{I}}
\newcommand{\mK}{\mathcal{K}}
\newcommand{\mP}{\mathcal{P}}
\newcommand{\mR}{\mathcal{R}}
\newcommand{\mU}{\mathcal{U}}
\newcommand{\mZ}{\mathcal{Z}}

\newcommand{\bA}{\mathbf{A}}
\newcommand{\bB}{\mathbf{B}}
\newcommand{\bC}{\mathbf{C}}
\newcommand{\bD}{\mathbf{D}}
\newcommand{\bF}{\mathbf{F}}
\newcommand{\btF}{\tilde{\mathbf{F}}}
\newcommand{\bG}{\mathbf{G}}
\newcommand{\bH}{\mathbf{H}}
\newcommand{\bI}{\mathbf{I}}
\newcommand{\bJ}{\mathbf{J}}
\newcommand{\bK}{\mathbf{K}}
\newcommand{\bL}{\mathbf{L}}
\newcommand{\bM}{\mathbf{M}}
\newcommand{\bN}{\mathbf{N}}
\newcommand{\bO}{\mathbf{O}}
\newcommand{\bP}{\mathbf{P}}
\newcommand{\btP}{\tilde{\mathbf{P}}}
\newcommand{\bQ}{\mathbf{Q}}
\newcommand{\bR}{\mathbf{R}}
\newcommand{\btR}{\tilde{\mathbf{R}}}
\newcommand{\btS}{\tilde{\mathbf{S}}}
\newcommand{\bS}{\mathbf{S}}
\newcommand{\bT}{\mathbf{T}}
\newcommand{\btT}{\tilde{\mathbf{T}}}
\newcommand{\bU}{\mathbf{U}}
\newcommand{\bV}{\mathbf{V}}
\newcommand{\btV}{\tilde{\mathbf{V}}}
\newcommand{\bW}{\mathbf{W}}
\newcommand{\bX}{\mathbf{X}}
\newcommand{\btX}{\tilde{\mathbf{X}}}
\newcommand{\bY}{\mathbf{Y}}
\newcommand{\btY}{\tilde{\mathbf{Y}}}
\newcommand{\bZ}{\mathbf{Z}}

\newcommand{\bzero}{\textbf{0}}
\newcommand{\bone}{\textbf{1}}
\newcommand{\ba}{\mathbf{a}}
\newcommand{\bb}{\mathbf{b}}
\newcommand{\bc}{\mathbf{c}}
\newcommand{\bd}{\mathbf{d}}
\newcommand{\be}{\mathbf{e}}
\newcommand{\bff}{\mathbf{f}}
\newcommand{\bg}{\mathbf{g}}
\newcommand{\bh}{\mathbf{h}}
\newcommand{\bi}{\mathbf{i}}
\newcommand{\bj}{\mathbf{j}}
\newcommand{\bl}{\mathbf{l}}
\newcommand{\bn}{\mathbf{n}}
\newcommand{\bo}{\mathbf{o}}
\newcommand{\bp}{\mathbf{p}}
\newcommand{\bq}{\mathbf{q}}
\newcommand{\br}{\mathbf{r}}
\newcommand{\bs}{\mathbf{s}}
\newcommand{\bts}{\tilde{\mathbf{s}}}
\newcommand{\bt}{\mathbf{t}}
\newcommand{\bu}{\mathbf{u}}
\newcommand{\btu}{\tilde{\mathbf{u}}}
\newcommand{\bv}{\mathbf{v}}
\newcommand{\btv}{\tilde{\mathbf{v}}}
\newcommand{\bbv}{\bar{\mathbf{v}}}
\newcommand{\bw}{\mathbf{w}}
\newcommand{\bx}{\mathbf{x}}
\newcommand{\btx}{\tilde{\mathbf{x}}}
\newcommand{\by}{\mathbf{y}}
\newcommand{\bty}{\tilde{\mathbf{y}}}
\newcommand{\bz}{\mathbf{z}}
\newcommand{\btz}{\tilde{\mathbf{z}}}
\newcommand{\bhz}{\hat{\mathbf{z}}}

\renewcommand{\vec}[1]{\boldsymbol{#1}}
\newcommand{\mat}[1]{\mathbf{#1}}
\newcommand{\set}[1]{\mathcal{#1}}

\newcommand{\real}[0]{\mathbb{R}}
\newcommand{\tb}[0]{\textbf}
\newcommand{\ti}[0]{\textit}
\newcommand{\et}[0]{\ti{et al.}}

\newcommand{\imageset}[0]{\set{I}}
\newcommand{\image}[0]{\mat{I}}

\newcommand{\poseset}[0]{\set{P}}
\newcommand{\transset}[0]{\set{T}}
\newcommand{\jointset}[0]{\set{J}}
\newcommand{\garmparamset}[0]{\set{G}}

\newcommand{\template}[0]{\mat{T}}
\newcommand{\garment}[0]{\mat{G}}
\newcommand{\blendweight}[0]{w}
\newcommand{\blendweights}[0]{\mat{W}}

\newcommand{\normal}[0]{{\mathbf{n}}}

\newcommand{\depth}[0]{\widehat{\vec{d}}}
\newcommand{\ankl}[0]{\mathbf{x}_{l}}
\newcommand{\ankr}[0]{\mathbf{x}_{r}}
\newcommand{\lplane}[0]{L_{p}}
\newcommand{\lamdp}[0]{\lambda_{p}}

\newcommand{\pose}[0]{\vec{\theta}}
\newcommand{\shape}[0]{\vec{\beta}}

\newcommand{\joints}[0]{\mat{J}}
\newcommand{\jointsTwoD}[0]{\tilde{\mat{J}}}

\newcommand{\rot}[0]{\vec{R}}

\newcommand{\scale}[0]{\mat{s}}
\newcommand{\trans}[0]{\vec{t}}

\newcommand{\nplane}[0]{\bf{\widehat{{n}}} }

\newcommand{\garmparam}[0]{\vec{z}}
\newcommand{\offsets}[0]{\mathbf{D}}

\newcommand{\cut}{\mat{z}_\mathrm{cut}}
\newcommand{\style}{\mat{z}_\mathrm{style}}
\newcommand{\posenc}{\mat{P}_{\shape}}
\newcommand{\zpose}{\mat{z}_{\pose}}

\newcommand{\smpl}[0]{M}
\newcommand{\posefun}[0]{T}
\newcommand{\blendfun}[0]{W}
\newcommand{\offsetfun}[0]{B}
\newcommand{\offsetsfun}[0]{D}
\newcommand{\jointfun}[0]{J}
\newcommand{\garmfun}[0]{G}

\newcommand{\metricdist}{pairwise normalized distances between persons}

\newcommand{\numberOfMetrics}{three} 

\newcommand{\Modelname}{Keep Your Feet on the Ground} 

\newcommand{\lossRep}[0]{L}
\newcommand{\multilossRep}[0]{\hat{L}}
\newcommand{\lambdaRep}[0]{\lambda_{2D}}

\newcommand{\heightDistMetric}[0]{h_{err}}

\newcommand{\posevect}[0]{\vec{p}}
\newcommand{\posevectRef}[0]{\vec{q}}
\newcommand{\jointsDims}[0]{\mathbb{R}^{J\times3}}
\newcommand{\diffVect}[0]{\vec{\delta}}

\newcommand{\featureVect}[0]{\vec{f}}
\newcommand{\featureNet}[0]{\mathcal{G}}
\newcommand{\ourNet}[0]{\Phi}
\newcommand{\rfineNet}[0]{\mathcal{R}}

\newcommand{\reals}[0]{\mathbb{R}}

\newcommand{\xmark}{\ding{55}}%

\newcommand{\projpage}{\href{https://nicolasugrinovic.github.io/purposer/}{project page}}%

\newcommand{\red}[1]{{\color{red} {\bf #1}}}
\newcommand{\blue}[1]{{\color{blue} { #1}}}

\newcommand{\francescrmk}[1]{{\color{orange} {\bf FM: #1}}}
\newcommand{\francesc}[1]{\textcolor{orange}{#1}}
\newcommand{\gregrmk}[1]{{\color{olive} {\bf GR: #1}}}
\newcommand{\greg}[1]{\textcolor{olive}{#1}}
\newcommand{\philrmk}[1]{{\color{CornflowerBlue} {\bf PW: #1}}}
\newcommand{\fabrmk}[1]{{\color{teal} {\bf FB: #1}}}
\newcommand{\thomasrmk}[1]{{\color{red} {\bf TL: #1}}}
\newcommand{\nurmk}[1]{{\color{Purple} {\bf NU: #1}}}
\newcommand{\tl}[1]{{\color{BrickRed} {#1}}}
\newcommand{\nuk}[1]{{\color{Purple} {#1}}}
\newcommand{\pw}[1]{\textcolor{CornflowerBlue}{#1}}
\newcommand{\fb}[1]{\textcolor{teal}{#1}}
\newcommand{\nukr}[1]{{\color{Purple} {\bf #1}}}
\newcommand{\todo}[1]{{\color{red}[TODO: #1]}}

\newcommand{\gray}[1]{{\color{Gray} {\bf #1}}}

\newcommand{\purple}[1]{{\color{Purple} {#1}}}
\newcommand{\green}[1]{{\color{OliveGreen} {#1}}}

\def\paperID{128} 
\def\confName{3DV\xspace}
\def\confYear{2024\xspace}

\title{Purposer: Putting Human Motion Generation in Context}

\author{
Nicolas Ugrinovic${}^1$
\hspace{0.4cm}Thomas Lucas${}^2$\hspace{0.4cm}Fabien Baradel${}^2$\hspace{0.4cm}\hspace{0.4cm}Philippe Weinzaepfel${}^2$\\Gr\'egory Rogez${}^2$\hspace{0.4cm} \hspace{0.4cm} Francesc Moreno-Noguer${}^1$ 
\and
${}^1$Institut de Robòtica i Informàtica Industrial, CSIC-UPC, Barcelona, Spain\\
${}^2$NAVER LABS Europe\\
}

\twocolumn[{%
\renewcommand\twocolumn[1][]{#1} %
\maketitle
\thispagestyle{empty}
\vspace{-0.75cm}
\begin{center}
    \centering
    \includegraphics[width=1.0\linewidth, trim={0cm 0.2cm 0cm 0.2cm}, clip = true]{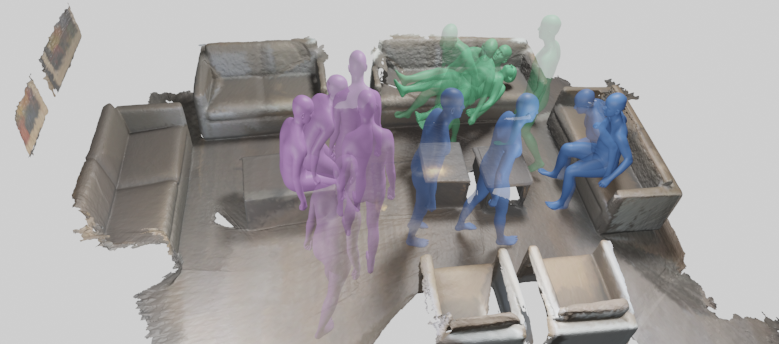} \\[-0.3cm]
    \captionof{figure}{\small{\textbf{An example of human motion generation in context.} We propose a method able to generate realistic-looking motions that interact with virtual scenes. In this example we take a scene from ScanNet~\cite{dai2017scannet}. The motion can be controlled with semantic action/object queries: here the human is first commanded `\textit{\purple{sit on table}}', then \blue{`\textit{sit on couch}'}, and finally \green{`\textit{lie on couch}'}. 
    \ours is a learning-based probabilistic model that can work efficiently with diverse types of conditioning. 
    }
    }\label{fig:teaser_main}
    \vspace{0.22cm}
\end{center}
}]
\begin{abstract}
     We present a novel method to generate human motion to populate 3D indoor scenes.
     It can be controlled with various combinations of conditioning signals such as a path in a scene, target poses, past motions, and scenes represented as 3D point clouds.
    State-of-the-art methods are either models specialized to one single setting, require vast amounts of high-quality and diverse training data, or are unconditional models that do not integrate scene or other contextual information. As a consequence, they have limited applicability and rely on costly training data. To address these limitations, we propose a new method ,dubbed \ours, based on neural discrete representation learning. Our model is capable of exploiting, in a flexible manner, different types of information already present in open access large-scale datasets such as AMASS. First, we encode unconditional human motion into a discrete latent space. Second, an auto-regressive generative model, conditioned with key contextual information, either with prompting or additive tokens, and trained for next-step prediction in this space, synthesizes sequences of latent indices. We further design a novel conditioning block to handle future conditioning information in such a causal model by using a network with two branches to compute separate stacks of features. In this manner, \ours can generate realistic motion sequences in diverse test scenes. Through exhaustive evaluation, we demonstrate that our multi-contextual solution outperforms existing specialized approaches for specific contextual information, both in terms of quality and diversity. Our model is trained with short sequences, but a byproduct of being able to use various conditioning signals is that at test time different combinations can be used to chain short sequences together and generate long motions within a context scene. 
    \vspace{-0.3cm}
\end{abstract}
\section{Introduction}
\label{sec:intro}
Generating realistic and diverse human motion is a decades-old research problem \cite{Badler1975thesis,Badler1993} but has gained traction in recent years \cite{actor,temos,teach,posegpt,gamma:cvpr22}. In this work, we propose a learning-based model for motion generation that can be controlled using various forms of \emph{contextual information} in order to navigate and interact with virtual scenes. In practice and as illustrated in Figure~\ref{fig:teaser_main}, human motion is strongly determined by several forms of context. Among them are: scene geometry, semantics of the surrounding objects, past motion and target actions and poses. So far, already established approaches have focused on narrow subsets of these. For instance, \cite{posegpt,mao2022weakly} both condition on actions and past poses but do not consider scene or target goals. 
To extend their applicability to VR/AR and other potential areas, the generated motion needs to make sense for a given scene. This requires taking into account past motion \cite{Habibie2017ARV,BarsoumCVPRW2018,aksan2019structured,yuan2020dlow,zhang2021we} together with scene geometry \cite{wang2020synthesizing,hassan_samp_2021,Wang_2022_CVPR}.
However, human motion data in context is scarce; this hinders the development of powerful conditional models. 

In the PROX~\cite{PROX:2019} dataset, the amount of human motion data available together with detailed scene information is two orders of magnitude smaller than AMASS~\cite{AMASS:ICCV:2019}. The lack of conditional data limits the expressivity of the models used and is not the regime in which recent deep learning methods excel.
In that scarce data regime, existing scene-conditioned methods rely on test time optimization loops, which allow them to effectively take into account scene boundaries, but affects the realism of the generated motion~\cite{wang2020synthesizing,Wang_2022_CVPR}. In contrast, we leverage the recent HUMANISE dataset~\cite{wang2022humanise} to learn scene interactions from the data. We also use unconditional data from where we learn a powerful motion prior.

We build our model on top of PoseGPT~\cite{posegpt}, itself based on neural discrete representation learning \cite{vqvae}. Thus, in our model, human motion is first mapped into an abstract discrete feature space, \emph{without any conditioning}. Any human motion given as input can be represented as a \emph{trajectory} in that discrete latent space, \ie, a sequence of centroids. After this, motion is modeled in a \emph{probabilistic} manner, directly in that latent space, by predicting latent trajectories in an auto-regressive manner. At this stage, various forms of contextual information can be used to condition the model and reduce prediction uncertainty. The latent trajectory is then mapped back into a continuous motion representation and latent trajectories are finally decoded into motion.

We propose a method that can take advantage of various combinations of contextual information. We account for three broad categories of contextual information, that can be combined together arbitrarily. First, we use the \textbf{scene geometry.} The scene is represented as a point cloud, encoded, and used to condition our generative model in latent space to exploit this information. Second, we use \textbf{past observations and future targets.} A limitation in existing auto-regressive approaches is that they cannot easily be conditioned on time-dependent future information, because of their causal design. To remedy this, we propose a simple and flexible architecture that allows us to effectively condition our model on future trajectories or randomly selected future poses. Finally, we use \textbf{semantic information.} To achieve semantic control, we condition the second stage model on \emph{target poses} which are generated using pairs of actions and object labels as targets as proposed in ~\cite{Zhao:ECCV:2022}. This offers semantic control over the generated sequences.

By combining this with conditioning on the past, we are able to chain multiple action/object targets together, which offers even more flexible semantic control and allows us to generate longer motion sequences, despite training on short-term sequences (HUMANISE). For instance, one can generate long sequences with multiple actions at different locations in the scene (\eg conditioned on an interaction with nearby objects) while using a conditioning corresponding to locomotion to navigate  (\ie, move along a path in the scene from this first object to this second object).

In summary, we present a model capable of leveraging unconditional data together with combinations of contextual information and generate motion to populate virtual scenes. Our model (a) can leverage large amounts of unconditional data, (b) can adapt to various contexts and (c) offers fine control on model outputs.

We train our auto-encoder on large-scale unconditional data from the BABEL dataset~\cite{babel} and our auto-regressive component with various combinations of conditioning signals on the HUMANISE dataset~\cite{wang2022humanise} and further fine-tune it on PROX~\cite{PROX:2019}. To evaluate our approach, we measure sample quality and sample diversity, as well as our model's generalization capabilities following practices established by existing work on uncontextualized motion generation~\cite{actor,posegpt}, inspired from the image generative modeling literature~\cite{naeem2020reliable,barratt2018note,shmelkov2018good,adeLucas}.
We also evaluate the synthesized motion's coherence with the scene, namely \textit{physical plausibility}, using contact and non-collision scores~\cite{wang2020synthesizing,Zhang:PSI:CVPR:2019}. In this manner, we show that our proposed approach generates high-quality motions to populate virtual scenes. We provide video results and code at the \projpage.

\section{Related work}
\label{sec:related_work}
\noindent \textbf{Human motion generation.}
The task of class-conditional human motion synthesis was first tackled assuming cyclic human actions such as walking \cite{urtasun2007modeling,taylor2006modeling}.
 More recent work have focused on adapting generative models to action conditional 3D human motion generation~\cite{actor,chuan2020action2motion}, and some approaches have explored conditioning on past poses~\cite{posegpt,mao2022weakly}. However, these methods do not condition on contextual information about the scene, which limits their applicability in practice. Another promising research avenue to control generated motion is to condition the model on high-level but detailed textual descriptions, as explored in \cite{lin2018human,ahn2018text2action,companim,animlang,temos,teach,motionClip,t2mgpt,tevet2023hu_mo_diff_mo,shafir2023human} or audio representations ~\cite{li2021learn,lee2019dancing}. 
 While these approaches offer fine-grained control over the generated motions, they do not allow to generate motions in a given environment.

\noindent \textbf{Scene interaction synthesis.}
It was not until recent years that the community focused its attention on estimating~\cite{PROX:2019,Corona_cvpr2020b,ugrinovic2021body} and generating~\cite{Zhang:PSI:CVPR:2019,PLACE:3DV:2020,Hassan:POSA:CVPR:2021} human poses taking into account a 3D scene context. This was shown to improve both 3D pose and motion estimations~\cite{PROX:2019,humor,zhang2021_lemo,shimada2022hulc,Luo2022EmbodiedSH}.  
Most recently, COINS~\cite{Zhao:ECCV:2022} propose a framework that adds semantic control to this generation process. By augmenting the PROX dataset~\cite{PROX:2019} with action-object paired labels and developing a specialized model, they generate semantically coherent poses. 
Building on their work, we go beyond static poses and propose a \emph{motion} model that can be conditioned on action object pairs.

\noindent \textbf{Object-conditioned human motion generation.}
One existing line of work focuses on conditioning motion generation on contextual or interaction information, be it nearby small~\cite{taheri2021goal,wu2022saga}, medium~\cite{zhang2022couch} or dynamic~\cite{christen2022dgrasp} objects. 
In these cases, emphasis is given to one single object at a time. GOAL~\cite{taheri2021goal} and SAGA~\cite{wu2022saga} focus on generating whole-body motions to match a final hand-grasping pose. COUCH~\cite{zhang2022couch} on the other hand focuses only on chairs, thus capturing human-chair interactions. By contrast, we focus on modeling more general interactions between human motion and an unconstrained number of objects within a scene.  

\noindent \textbf{Scene-conditioned human motion generation.}
 Up until now, few works have fully studied scene-conditioned human motion generation ~\cite{starke2019nsm,wang2020synthesizing,hassan_samp_2021,gamma:cvpr22,Wang_2022_CVPR}. 
 Neural State Machine~\cite{starke2019nsm} generates different modes of motion that can be blended between different actions while interacting with the environment. 
 This model allows excellent motion control while providing smooth transitions between modes. However, this method was designed for simple hand-crafted environments and relies on a deterministic model, limiting its ability to produce diverse motion and to model the full extent of human motion.
 Recently,~\cite{wang2020synthesizing} tackles the task of generating long-term motion given a 3D scene and start/end goal positions using a hierarchical framework
that decomposes the task by synthesizing shorter motion sequences.
This method relies on a post-optimization step to ensure smoothness, robust foot contact, and avoiding collisions with the scene. While effective, this optimization step reduces the naturality of the motion.
SAMP~\cite{hassan_samp_2021} creates human motion conditioned on the action and a final target object, position and orientation in a stochastic manner.
Given a starting position, and a target object (\eg chair, sofa), they first estimate a goal position and orientation and then estimate a plausible path between the start and goal positions. Finally, they generate a sequence of poses with an auto-regressive conditional Variational Auto-Encoder (cVAE). 
Wang \et~\cite{Wang_2022_CVPR} propose a model composed of various networks each specialized on one sub-task: generating target start/end poses, path planning, and sequential human poses generation. They rely on cVAE networks conditioned on actions and on a generated path. However, they use the same optimization step as in~\cite{wang2020synthesizing} to reduce foot skating and scene penetration and thus suffer from a similar lack of naturality. 

Reinforcement learning methods have also been used to tackle this problem~\cite{gamma:cvpr22,rempeluo2023tracepace,hassan_siggraph23,dimos_iccv23}. Zhang~\et~\cite{gamma:cvpr22} and Rempe~\et~\cite{rempeluo2023tracepace} mostly focus on generating realistic locomotion taking the scene topology into account. Hassan~\et~\cite{hassan_siggraph23} use a physical simulated character and imitation learning to generate diverse actions within an environment. Concurrent to our work, DIMOS~\cite{dimos_iccv23} extends~\cite{gamma:cvpr22} to include more actions that interact with the environment.  

Finally, \cite{wang2022humanise} contributed a synthetic dataset, HUMANISE, that places a subset of motion capture (MoCap) sequences from AMASS~\cite{AMASS:ICCV:2019} dataset in scenes from~\cite{dai2017scannet}. In this work, we leverage HUMANISE dataset to include scene context. This way, we are able to generate realistic motions of humans navigating and interacting in a scene. Our approach is most similar to~\cite{wang2020synthesizing,Wang_2022_CVPR}, but yield richer interactions and more realistic motions. We present a direct comparison to these two approaches in Section~\ref{sec:experiments}.

\newcommand{\action}{{\mathbf{a}}}
\newcommand{\quantizedcode}{z_{\mathbf{q}}}
\newcommand{\codebookdim}{N_z}
\newcommand{\RR}{\mathbb{R}}
\newcommand{\decoder}{D}
\newcommand{\encoder}{E}
\newcommand{\gpt}{G}
\newcommand{\hz}{\hat{\bm{z}}}
\newcommand{\quantize}{Q}
\newcommand{\codebook}{\mathcal{Z}}
\newcommand{\Attn}{\operatorname{Attn}}
\newcommand{\softmax}{\operatorname{softmax}}

\section{Purposer}
We build our \ours model on auto-regressive discrete-based generative models such as PoseGPT~\cite{posegpt}, T2M2~\cite{Tm2t}, T2M-GPT~\cite{t2mgpt} or Bailando~\cite{bailando}.
we detail how we propose to condition such causal methods, in particular in the case of future conditioning, in Section~\ref{sec:method:conditioning}. 
We then discuss the different forms of contextual information that we consider for motion generation (Section~\ref{sec:method:context}). 
Finally, we detail our training setup in Section~\ref{sec:method:training} and how various conditioning signals can be combined to generate long-term sequences while being trained on short ones (Section~\ref{sec:method:longterm}).

\subsection{Background on discrete auto-regressive models}
\label{sec:method:background}

\begin{figure}[h!]
\centering
\includegraphics[width=1\linewidth]{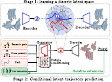} \\[-0.2cm]
\caption{
\textbf{Method Overview.} An auto-encoder is learned to compress human motion, without any context, into a discrete latent sequence space (top). A probabilistic model (bottom) is trained directly in that space, with three types of optional context: (a) scene geometry, (b) semantic goals, (c) observation of past motion. 
\vspace*{-0.3cm}
}
\label{fig:overview}
\end{figure}

Discretization-based auto-regressive models proceed in two stages, see Figure~\ref{fig:overview}: (a) an auto-encoder is learned to move from the continuous input space to a discrete latent space and vice-versa, (b) an auto-regressive model is learned in this discrete space, and can be fed to the decoder for obtaining the output in the desired space. We now give more background on these two stages.

\noindent\textbf{Discrete motion auto-encoder.}
An auto-encoder is learned to compress motion sequences into discrete latent representations with neural discrete representation learning~\cite{vqvae}, see top row of Figure~\ref{fig:overview}. Concretely, an encoder $\encoder(\cdot)$, a quantizer  $\quantize(\cdot)$ with a codebook and a decoder $\decoder(\cdot)$ are trained such that the reconstruction error is minimized. A given motion sequence $\bp$ of length $T$ can be represented by a discrete sequence of indices $\bz = \{ \bz_1, \hdots, \bz_{T'} \}$  of length $T^{'}$,
by computing $\quantize(\encoder(\bp))$. Conversely, any sequence of discrete latent indices can be decoded into a motion sequence by forwarding it to the decoder $\decoder$. Note that here, we use $T^{'}$ instead of $T$ as the sequence in pose space $\bp$ can be downsampled when converting to the latent discrete space $\bz$ and then upsampled again by using the decoder $\decoder$. 
To allow conditioning on past observations, a causal encoder is used, such that for any $t \le T^{'}$, $\bhz_t$ is a function of $\{\bp_1,\ldots,\bp_{\floor{t \cdot T/T^{'}}}\}$ only.
In this work, we rely on the discrete motion auto-encoder from PoseGPT~\cite{posegpt} that further uses product quantization for better leveraging the discrete space.

\begin{figure*}[t!]
\centering
\includegraphics[width=0.9\linewidth,trim={0cm 0.0cm 0cm 0.0cm},clip]{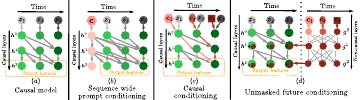} \\[-0.3cm]
\caption{\label{fig:training_future}\textbf{Ways of conditioning an auto-regressive model.} \textit{(a)}: an auto-regressive model without conditioning is based on causal attention. \textit{(b)}: by adding a prompt token  $ \bm{c}_0$ to the sequence, sequence-wide conditioning can be added. \textit{(c)}: for time-dependent conditioning $\bm{c}_1,\hdots,\bm{c}_{T'}$, features could be combined but the model will be unaware of the future conditioning when predicting a given timestep. \textit{(d)}: we include future conditioning by making a non-causal network to process the time-varying conditioning, and combine their future with the standard causal generative model. 
\vspace*{-0.2cm}
}
\end{figure*}

\noindent\textbf{Auto-regressive prediction.}
The auto-regressive model can then be learned directly in the frozen discretized latent space.
Any motion sequence $\bp$ of length $T$ can be represented as $\bz = \{ \bz_1, \hdots, \bz_{T'} \}$.
To generate trajectories in the latent space, an auto-regressive model $G(\cdot)$ can be trained to predict the next index as the successful GPT~\cite{gpt} family in natural language processing, \ie, by maximizing: 
\begin{equation}
\label{eq:gpt}
p_{G}(\bz) = p\left(\bz_1\right)\prod_{t=2}^{T^{'}} p\left(\bz_{t} | \bz_{1}, \hdots, \bz_{t-1}\right).
\end{equation} 
To obtain motion samples, latent sequences are sampled from $p_{G}$ and decoded using the decoder $\decoder(\cdot)$.
Such auto-regressive models can elegantly be conditioned on past motion when using a causal encoder.

\subsection{Future conditioning in auto-regressive models}
\label{sec:method:conditioning}
While conditioning an auto-regressive model such as a GPT with a \textit{sequence-wide} information, \ie, a fixed context across the full sequence (\eg, static scene information), can be easily implemented, it is not straightforward to condition 
on \textit{future} information (\eg, a target pose or a path). 
\noindent \textbf{Sequence-wide conditioning.} 
Some types of conditioning are valid for the full sequence -- for instance static scene information, a sequence duration $T$, or a constant action label. In that case, given an input sequence $(\bz_1, \hdots, \bz_{T^{'}})$ embedded into features $\bm{h} = (\bm{h}_1, \hdots, \bm{h}_{T^{'}})$ and some conditioning signal $\bm{c}$ represented by a feature vector $\bm{h}_c$, conditioning the auto-regressive model can be done simply by \emph{prompting}, \ie, adding $\bm{h}_{\bm{c}}$ as an extra token at the start of the input sequence, as commonly done \eg in language models~\cite{ouyang2022training_lang_to_follow}:
\begin{equation}
    \bm{\tilde{h}}_{\text{prompt}} = (\bm{h}_{\bm{c}}, \bm{h}_1, \hdots, \bm{h}_{T^{'}}).
\label{eq:prompt}
\end{equation}
Another solution is to inject it into all input tokens:
\begin{equation}
    \bm{\tilde{h}}_{\text{feat}} = (\bm{h}_1 \oplus \bm{h}_{\bm{c}}, , \hdots, \bm{h}_{T^{'}} \oplus \bm{h}_{\bm{c}} ),
\label{eq:feat}
\end{equation}
where the $\oplus$ operation denotes any operator that combines the two features, such as concatenation or sum.

\noindent \textbf{Conditioning with causal masking.}
Let us now consider a time-dependent conditioning $\bm{c}_1,\hdots,\bm{c}_{T^{'}}$ , \ie, an information from the `future'; for instance, the path to be followed defined as set of locations that varies with $t$.
Conditioning the input features directly as in Equation~\ref{eq:feat} remains possible, by replacing $\bm{h}_{\bm{c}}$ by $\bm{h}_{\bm{c}_t}$ at each timestep $t$. However, because of the causal masking, an auto-regressive model predicts $\bz_i$ only from past information; therefore with the conditioning in Equation~\ref{eq:feat} only past context is available when predicting a timestep $i$. The model then has to predict the future path rather than use the available information, which will deteriorate output quality.
This is probably why most methods are limited to a sequence-wide context, \ie, a single action label for~\cite{posegpt} or a text prompt represented with CLIP features~\cite{t2mgpt}. We now detail our proposed solution to circumvent this issue.

\noindent \textbf{Future conditioning.}
To implement future conditioning, we propose to use a network with two branches to compute two stacks of features -- a causal one responsible for the prediction of the next timestep and a non-causal one that can propagate information about the conditioning at all timesteps -- and inject the non-causal one into the causal one. More precisely, given an input token sequence $(\bz_1, \hdots, \bz_{T^{'}})$ and a conditioning sequence $(\bm{c}_1, \hdots, \bm{c}_{T^{'}})$,
both are embedded into feature sequences $\bm{h}^0$ and $\bm{g}^0$, respectively.  

As in standard auto-regressive models, a stack of  $L$ \emph{causal} layers $\bm{f}^1_{c}, \hdots, \bm{f}^L_c$ is used to compute features $\bm{h}^1, \hdots, \bm{h}^L$ such that for any $l$ and any $t$, $\bm{h}^l_t$ is a function of $\bz_1, \hdots, \bz_{t-1}$ only. In addition, a second stack of non-causal layers $\bm{f}^1, \hdots, \bm{f}^L$ is used to process $\bm{g}^0$, and for any $1 \le t \le T^{'}$:  
\begin{equation}
    \left\{
                \begin{array}{ll}
                  \bm{h}^l_t = \bm{f}^l_c(h^{l-1}_1, \hdots, \bm{h}^{l-1}_{t-1}),\\
                  \bm{g}^l_t = \bm{f}^l(\bm{g}^{l-1}_1, \hdots, \bm{g}^{l-1}_{t}, \hdots, \bm{g}^{l-1}_T),\\
                  \bm{\tilde{h}}^l_t = \bm{h}^l_t + \bm{g}^l_t.
                \end{array}
              \right.
\label{eq:future}
\end{equation}
With this construction,
 $\bm{h}^L_t$ is a function of $\bz_1, \hdots, \bz_{t-1}$ only and can be used to predict $\bz_t$, see Figure~\ref{fig:training_future} for an illustration. 
 Additionally, any feature $\bm{h}^l_t$ is a function of all conditioning signals $(\bm{c}_1, \hdots, \bm{c}_{T^{'}})$, and increasingly complex conditioning features $\bm{g}^1, \hdots, \bm{g}^L$ can be learned.
 This construction circumvents the causal masking, and the standard auto-regressive architecture does not need other modifications.

\begin{figure}[t!]
\centering
\includegraphics[width=0.98\linewidth]{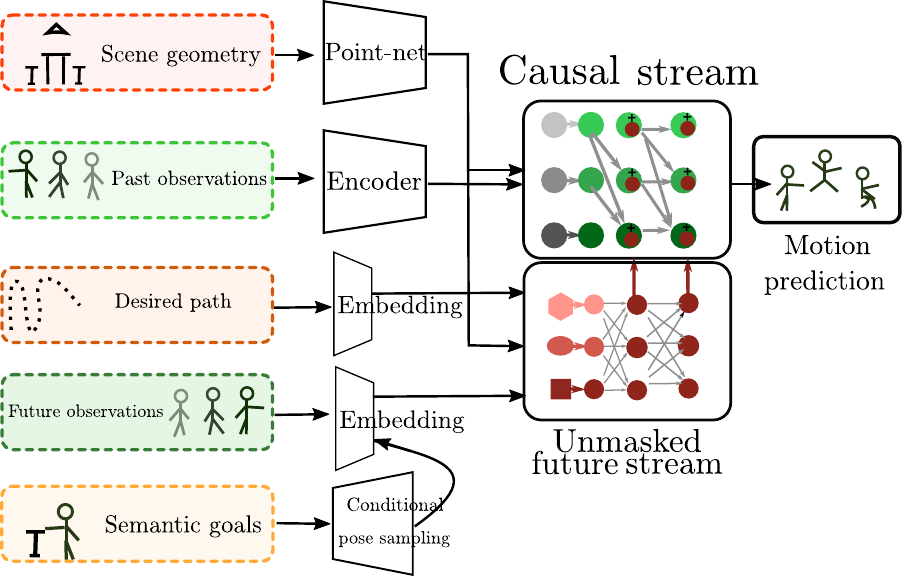} \\[-0.3cm]
\caption{\label{fig:conditioning}\textbf{Different conditionings used in \ours:}
high-level view of how different motion contexts are implemented.  \vspace*{-0.3cm} 
}

\end{figure}

\subsection{Motion generation with context}
\label{sec:method:context}

We now describe the different types of context we consider to put human motion generation in context, namely scene geometry, past observations or future target trajectories, and semantic information such as action labels or target human-object interaction.

\vspace{1mm}
\noindent\textbf{Conditioning on scene geometry.} 
To condition motions on a scene, we represent the geometry of the scene, given as an input point cloud, with a constant feature embedding $\bf{h}_c$ and condition with prompting.
A scene point cloud $\mathcal{S} = \{\bs_i | i = 1, \hdots N_s \}$ is embedded using PointNet~\cite{pointnet} following~\cite{wang2020synthesizing}.
The output is then projected with a learnable linear layer $\bW_s$:
$ \bc_s = \bW_s \cdot \texttt{PointNet}(\mathcal{S})$,
with $\bc_s$ the vector containing the scene information to condition  $p_G$ with prompting.

\vspace{1mm}
\noindent\textbf{Conditioning on past observations.}
An auto-regressive architecture naturally allows conditioning on observed past motion, as long as the latent sequence is produced by a causal encoder.
More precisely, if a past motion $\bp_1, \hdots, \bp_t$ of arbitrary length is observed as context, it can be encoded by the encoder into $\encoder(\bp_1, \hdots, \bp_t) = \bz_1, \hdots, \bz_{t'}$.
A future human motion of length $T$ can be sampled from our model conditioned on the observation: 
\begin{equation}
p_G(\bz|\bz_1, \hdots \bz_{t^{'}}) = \prod_{l=t^{'}+1}^{T^{'}+t^{'}} p(\bz_l | \bz_1, \hdots \bz_{l-1}).
\end{equation}
Thus by design, an auto-regressive model has the flexibility to be conditioned on past motion without change and/or retraining.

\vspace{1mm}
\noindent\textbf{Conditioning on trajectories and target poses.}
In addition to past observations, our model can also be conditioned on future target poses, or trajectories. For this type of conditioning, the \textit{causal conditioning} approach taken by PoseGPT does not allow the future path or targets to be observed when generating a given timestep. However, by using the \textit{future conditioning} from Section~\ref{sec:method:conditioning}, we can condition on trajectories or arbitrarily chosen future poses. Note that while this would allow us to condition the model on a future pose in arbitrary time steps in the future, for our purposes we chose the last pose in the sequence. Let $((x_1,y_1),\hdots,(x_{T'},y_{T'}))$ be a 2D trajectory on a bird's eye view of the scene, $\bp_{tj}$ some future poses, and $\bm{W}_p$ and $\bm{W}_f$  two learnable linear layers. We can write the path and future poses conditions as:
\begin{eqnarray}
\bm{c}_p &=& (\bm{W}_p \cdot (x_1, y_1), \hdots, \bm{W}_p \cdot (x_T, y_T))\\ 
\bm{c}_f &=& (\bm{W}_f \cdot \bm{p}_{t})_{t \in \mathcal{T}}\;
\end{eqnarray} 
where $\mathcal{T}$ is a set of time steps;
$\bm{c}_p$ and $\bm{c}_f$ are then used as input to compute $h^0$ in Equation~\ref{eq:future}.
Note that conditioning on the final target pose is a special case of this, which could also be achieved with a simpler conditioning such as an additional token. 

\vspace{1mm}
\noindent\textbf{Conditioning on semantic information.}
We consider two semantic contexts: (a) \emph{action labels} and (b) a \emph{target final human-object interaction}.

\vspace{1mm}
\noindent$\circ \hspace{0.1cm}$ \emph{Action labels:} We use sequence-wide conditioning to include action information into the model. 

Action labels are embedded and projected into $\bc_a$ and $p_G$ is conditioned on the result by inputting $\bc_a$ to the unmasked stream.

\vspace{1mm}
\noindent$\circ \hspace{0.1cm}$ \emph{Target human-object interactions:}  We also integrate the possibility to control the interaction of the motion with specific objects by conditioning our motion samples on \emph{pairs} of actions and objects, 
$\{(a_t, o_t)\}$, \eg (lie, couch). 
To achieve that, we decouple the problem into (a) generating a \emph{static} pose $\bp_{(a,o)}$ conditionally on $(a, o)$, following the pioneering approach of \cite{Zhao:ECCV:2022} ,  
and (b) conditioning the motion model on this target pose, which is embedded into $\bc_{(a,o)} = \bW_{ao} \cdot \bp_{(a,o)}$, where $\bW_{ao}$ is a learnable linear layer.
Then, a latent sequence is sampled using this pose as target, and the latent sequence is decoded into a human motion with the decoder $\decoder(\cdot)$.
At train time, we condition the model on final target poses extracted from the data rather than generated.

Finally, based on the different types of conditioning that we have introduced above we can re-write Equation (\ref{eq:gpt}) with the following conditioning:
\begin{equation}
p_{G}(\bz_i|\bc_i) = \prod_{t=1}^{T^{'}} p\Big(\bz_t | \{\bz_i\}_{i<t};  \bc_s, \bc_{a}, \bc_{(a,o),t}, \bc_p, \bc_f \Big).
\end{equation}
In Figure~\ref{fig:conditioning}, we recap how the different contextual information is embedded. To generate motion, latent variables are iteratively sampled and added to the conditioning sequence before being decoded.

\subsection{Training setup}
\label{sec:method:training}

We directly use the discrete auto-encoder from \cite{posegpt} trained on BABEL~\cite{babel}.
To train the auto-regressive model 
we use the scene-conditioned data from HUMANISE,
 a synthetic dataset composed of a subset of BABEL motions placed in ScanNet~\cite{dai2017scannet} scenes. It contains 19.6K human motion sequences in 643 3D scenes and consists of actions that are commonly performed when interacting with a scene. We train each or our generative networks with the relevant inputs as conditioning. Finally, to evaluate on real-world but smaller-scale data, we fine-tune the model on PROX~\cite{PROX:2019}, which consists of 100K frames with pseudo ground truth.
It captures dynamic sequences of 20 subjects in 12 scenes at 30fps. Implementation details are included in the supplementary material. 

\subsection{Generating long-term sequences}
\label{sec:method:longterm}

While our model is trained on short sequences, taking advantage of its autoregressive nature,  and using various sets of conditioning allows us to generate long-term motions that are coherent with a virtual scene by chaining short-term motions.  
Specifically, we can define an `object interaction' configuration to generate motions that interact with scene objects, \eg sit/lie down, by giving the proper conditioning as input. 
We can then use a `locomotion' configuration to generate walking motion sequences to navigate the scene and connect between different interaction motions that are far apart in the scene, if needed. 
Each of these configurations is conditioned with relevant information. The `object interaction' model is mainly conditioned with a target pose that encodes a correct interaction to guide the motion to reach that pose at the end of the sequence. Other than the target pose, this model is configured with the rest of the relevant conditioning: action labels, scene geometry and past observations.
The `locomotion' model is mainly guided with a path, \eg the shortest path between the two locations where the two motion interactions happen in the scene, \eg from the A$^*$ algorithm. Although, this path could be set in any other different manner if needed. The `locomotion' configuration does not need the use of a target pose but instead only the desired $(x,y)$ position is specified. 
For chaining any two consecutive sequences together we take the last $n$ poses from the first sequence and use those to condition the generation of the next sequence. Concretely, we use $n=2$ for our results as the donwsample factor from the pose space $\bp$ to the latent space $\bz$ is 2. Thus, the minimum number of conditioning poses is $n=2$ (Section~\ref{sec:method:background}).  
For more detail about the exact conditioning both of these configurations please refer to Table~\ref{tab:ablation} in Section~\ref{sec:experiments:ablations}.

\section{Experiments}
\label{sec:experiments}
 Given that without any conditioning our method boils down to a standard discrete auto-regressive model such as~\cite{posegpt}, we focus our experiments on different conditioning scenarios. 
 We aim to apply our model to populate virtual scenes. We thus focus our experiments on scene-conditioned motion generation, with various types of context on top. After introducing the datasets and metrics in Section~\ref{sec:experiments:metrics}, we present several ablations in Section~\ref{sec:experiments:ablations}. Finally, we compare our method to the state of the art in Section~\ref{sec:experiments:sota}. 

\noindent \textbf{Conditionings.} We clearly state in each table the conditioning that was used. In addition to scene conditioning either at the feature-level (F) or with prompting, \ie, an extra token (T), we consider other conditioning forms: first pose (first), target pose (target/P) or target position (target/XY), action label and path. Action labels conditioning is used in all cases unless noted otherwise.

\subsection{Datasets and metrics}
\label{sec:experiments:metrics}

\noindent \textbf{Datasets.} 
We perform most experiments on the HUMANISE dataset, following the official splits~\cite{wang2022humanise} with 16.5K motions in 543 scenes for training and 3.1K motions in 100 scenes for testing.
We also experiment on the PROX dataset with the standard splits (8 training scenes and 4 scenes for testing) as in~\cite{wang2020synthesizing,Wang_2022_CVPR}, and rely on the improved fittings provided in~\cite{zhang2021_lemo} and action labels from~\cite{Zhao:ECCV:2022}. 

\noindent\textbf{Physical plausibility.} 
We evaluate the physical plausibility of generated interactions using the non-collision metric proposed in ~\cite{Zhang:PSI:CVPR:2019}, which measures how much the generated human mesh interpenetrates the mesh of the scene, and the contact score proposed in~\cite{wang2020synthesizing} which is complementary as it ensures that the motion actually makes contact with the object -- non-collision alone would be maximized by standing away from everything. For the contact scores, we follow~\cite{wang2022humanise} and use a threshold of 0.02, except for some tables that are clearly specified where we follow~\cite{Wang_2022_CVPR,wang2020synthesizing} use 0.01. 

\noindent \textbf{Diversity metrics.} 
To evaluate the diversity of generated samples, we follow common practices in the literature~\cite{wang2020synthesizing,wang2022humanise,scene_diff_cvpr23} and report the average pairwise distance (APD) metric. This metric measures the average $L_2$ distance between all pairs of motion of $K$ samples computed with exactly the same input information. When evaluating on HUMANISE, we follow the practice of~\cite{wang2022humanise} .When comparing on PROX, we follow~\cite{scene_diff_cvpr23}. In both cases $K=20$.
Additionally, following~\cite{scene_diff_cvpr23}, we measure APD for a specific set of 61 markers (APD mark.) extracted from the body meshes.
As advocated for in \cite{posegpt}, and given that all components of our model are likelihood based, we also report likelihood-based metrics for the generator~\gpt. 

\noindent\textbf{Quality metrics.} 
To measure the quality of sampled motion, we compute the Fr\'echet distance score~\cite{fid}.
For comparison with existing work, we compute the FD metric with a VPoser~\cite{smplx} model.
 We denote this metric by FD$_\text{static}$ as this model only takes individual frames as input. Please also find qualitative video results in the supplementary for a complementary perspective.

\subsection{Ablation of design choices}
\label{sec:experiments:ablations}
In Table~\ref{tab:ablation}, we first ablate the impact of the different conditioning information used by our model: future stream, first and target poses, scene point-cloud, and path.  Note that `future stream' is not a type of information but a novel component of our model that conditions the information in a special way. The first two rows correspond to our baseline, namely, PoseGPT~\cite{posegpt}.

\noindent \textbf{First pose.} In Table~\ref{tab:ablation}, rows 1 and 2, we observe that using the first observed pose as conditioning significantly improves the non-collision and contact metrics. 
This is expected as it guides the motion in a correct direction where less collisions are likely to occur, weather the model is conditioned on the scene or not. Thus adding a first pose, which can be obtained from past observed motion, improves scene interaction and 
motion control
while retaining generation diversity.

\noindent \textbf{Scene.} Using scene information helps to improve the quality of the generated motion by forcing it to better fit the given scene. However, this depends on how we input this information to our model. As seen in Table~\ref{tab:ablation} row 3, if done at the feature level (F), \ie, by concatenation with the input embeddings, the model's performance does not improve, or even deteriorates, both in terms of next index prediction negative likelihood (NLL) and non-collision. On the other hand, if we introduce this information with token prompting (T) (row 4), we both maintain the generation quality (NLL) and improve penetration (69.56\% vs. 70.19\%).

\begin{table}[]
\centering
\resizebox{\linewidth}{!}{%
\begin{tabular}{c cccHc cccc}
\toprule
\multicolumn{6}{c}{conditioning} & NLL$\downarrow$ & APD$\uparrow$ & \multicolumn{2}{c}{phys. plausibility$\uparrow$} \\
stream & first & scene & target & posit. & path &  & mark. & non-coll.& contact \\
\cmidrule(lr){1-6} \cmidrule(lr){7-7}  \cmidrule(lr){8-8} \cmidrule(lr){9-10} 
\mCross    & \mCross    & \mCross    & \mCross    & \mCross    & \mCross    & 0.86          & 4.83          & 55.73          & 93.93          \\
\mCross    & \checkmark & \mCross    & \mCross    & \mCross    & \mCross    & 0.86          & 4.09          & 69.56          & 92.68          \\
\mCross    & \checkmark & F          & \mCross    & \mCross    & \mCross    & 0.96          & 4.08          & 69.10          & 93.10          \\
\mCross    & \checkmark & T          & \mCross    & \mCross    & \mCross    & 0.87          & 3.91          & 70.19          & 92.79          \\
\mCross    & \checkmark & T          & P          & \mCross    & \mCross    & 0.62          & 3.05          & 71.64          & 91.86          \\
\midrule
\checkmark & \checkmark & T          & XY        & \checkmark & \checkmark & 0.70          & \textbf{5.91} & 71.24          & 92.59          \\
\checkmark & \checkmark & T          & P          & \mCross & \checkmark    & 0.48          & 3.02          & 71.76          & 94.15          \\
\checkmark & \checkmark & T          & P          & \mCross    & \mCross    & \textbf{0.42} & 3.13          & \textbf{73.28} & \textbf{94.29}\\
\bottomrule
\end{tabular}
} \\[-0.2cm]
\caption{\textbf{Ablation study} on HUMANISE~\cite{wang2022humanise} with action label conditioning and without post-processing optimization. 
XY means that it uses target position instead of target pose (P). 
\vspace{-0.2cm}
}
\label{tab:ablation}
\end{table}

\noindent \textbf{Target pose.} To guide motion towards human-scene interaction, we design our model to take as input a target conditioning pose. These target poses can be taken from the ground-truth data or sampled at test time, given action object pairs. 
As observed in Table~\ref{tab:ablation} (row 5), using a target pose gives an important increase of 1.46\% in the non-collision score while also improving the generation quality (NLL). Contact score is slightly reduced, possibly as aconsequence of reducing scene penetration. Furthermore, APD decreased but this is expected, as the generated motion is more constrained and therefore there is less variability in the outputs.

\noindent \textbf{Future Stream.} 
The future conditioning block proposed in Section~\ref{sec:method:context} allows conditioning on time-step dependent future signal and is a key component of our method.
The last three rows of Table~\ref{tab:ablation} show that variants using this component outperforms non-`future stream' counterparts. We hypothesize that this is because (a) it enables the use of all available path information and (b) it is a more flexible way to condition on the rest of features, \eg, target pose and scene.  
Using the time-dependent path as input (row 6) yields an improvement in non-collision score (0.70\%), maintains contact score, and substantially improves NLL. 

We refer to this setting as the \emph{locomotion model}.
 The last row presents the performance when conditioned on a target pose, but no path. It obtains the best performance in motion quality, non-collision, and contact, and is referred to as the \emph{object interaction} model.
 Note that all configurations in Table~\ref{tab:ablation} can generate any action present in the dataset; thus, we evaluate them all together.

\begin{table}[]
\centering
\resizebox{\linewidth}{!}{%
\begin{tabular}{HHHHHccHcHHccHrH}
\toprule

Name                                 & act        & first      & last       & path       & \multirow{2}{*}{optim}    &NLL$\downarrow$  & APD (marker)  & APD$\uparrow$     & APD (transl)  & APD (params)  & FD${_{\text{static}} \downarrow}$ & 
\multicolumn{4}{c}{phys. plausibility$\uparrow$} \\
 &  &  &  &  &  &  & & (all) & & & & non-coll. & & contact&  \\
\cmidrule(lr){1-6} \cmidrule(lr){7-7} \cmidrule(lr){8-11} \cmidrule(lr){12-12} \cmidrule(lr){13-16}
ours & \mCross  & \checkmark & \checkmark & \mCross  & \mCross  & \textbf{2.69} & 1.40  & 2.06  & 1.56  & 0.85    & 30.11   & 95.23   & 89.82 & \textbf{99.98} & 99.98    \\
ours & \mCross  & \checkmark & \checkmark & \mCross  & \checkmark & \textbf{2.69} & 1.52  & \bf{3.02}  & 1.57  & 2.02  & \bf{29.76}  & \bf{99.24} & 73.69 & 99.96  & 99.99   \\
\bottomrule
\end{tabular}
}
\vspace{-0.25cm}
\caption{\textbf{Impact of the optimization step} on PROX, performed with first/last pose cond., scene prompt, but without action and path information to match the conditions of~\cite{wang2020synthesizing}. Contact score threshold of 0.01. 
\vspace*{-0.1cm}
}
\label{tab:sota:prox}
\end{table}

\noindent \textbf{Impact of post-processing optimization.}
Previous methods~\cite{wang2020synthesizing,Wang_2022_CVPR} leverage a post-processing step that optimizes the generated motions to avoid collision and favor contacts. We measure the impact of such post-processing optimization on the PROX dataset and later compare to other SOTA methods that apply the same post-processing, see Table~\ref{tab:sota:prox}. This optimization step slightly improves the physical plausibility scores, but produces less natural and stiffer motion. This is best appreciated in the qualitative results.

\begin{figure}[t]
\centering
\includegraphics[width=1\linewidth]{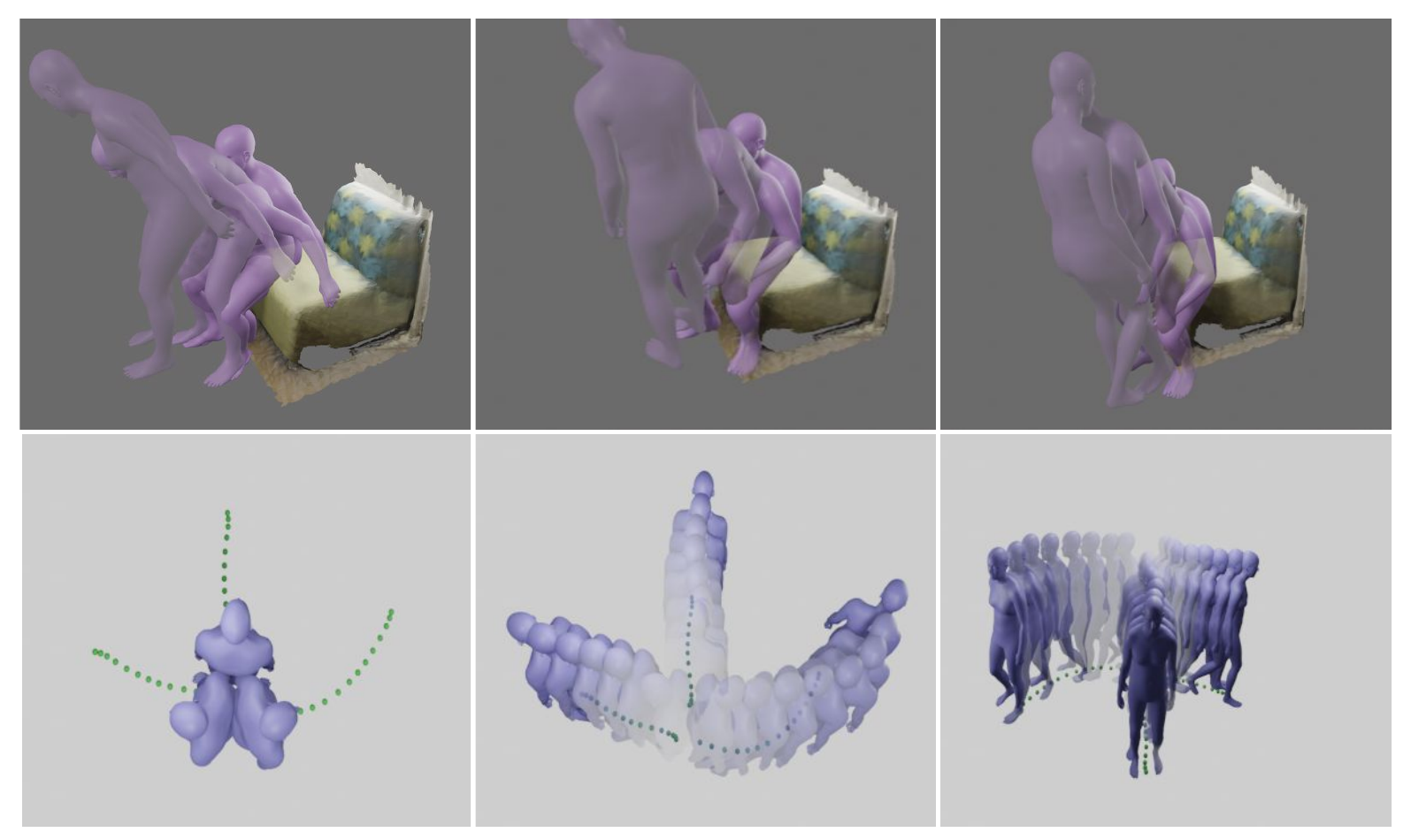} \\[-0.2cm]
\caption{
\textbf{Effect of target pose and path conditioning.}  \textbf{Upper row:} examples of object interaction. Here we use the same object with different and random initial body position and orientation. \textbf{Lower row:} demonstration of the effects of path conditioning: we can define the final position and trajectory given a common starting point. 
The green dots represent the conditioning path. 
}
\vspace{-0.4cm}
\label{fig:qualitatives}
\end{figure}

\noindent \textbf{Qualitative results and long-term generation.} 
In Figure~\ref{fig:qualitatives}, we present qualitative results that illustrate the impact of incorporating target pose conditioning (top row) to enable scene interaction. Our model is able to interact with the same object when initialized with random initial body locations and orientations. Additionally, we present the effect of the path conditioning (bottom row) for walking actions. Our model is capable of generating realistic locomotion in random directions and with arbitrarily chosen paths.
In the supplementary video, we also show examples of long-term motions combining short-term motions for object interaction and for locomotion, see Section~\ref{sec:method:longterm}.

\subsection{Comparison to the state of the art}
\label{sec:experiments:sota}
\begin{table}[]
\centering
\resizebox{0.5\textwidth}{!}{%
\begin{tabular}{l cHHcH HcHH c cHcH}
\toprule
Name                                 & \multicolumn{5}{c}{conditioning}                         & \multicolumn{4}{c}{APD$\uparrow$}  & FD${_{\text{static}}\downarrow}$ & \multicolumn{4}{c}{phys. plausibility$\uparrow$}         \\
                                 & act        & first      & last       & path      & optim      & mark. & (all) & tr. & par.             &                                & non-coll.    & thr=0 & contact       & thr=0.02         \\

\cmidrule(lr){1-1} \cmidrule(lr){2-6} \cmidrule(lr){7-10}  \cmidrule(lr){11-11}  \cmidrule(lr){12-15}
Wang \et \cite{wang2020synthesizing} & \mCross    & \checkmark & \checkmark & \mCross    & \checkmark & 0.00          & 0.00          & 0.00          & 0.00          & ---                                      & \textbf{99.91} & ---   & 99.35           & ---              \\
\bf{\ours}                                 & \mCross    & \checkmark & \checkmark & \mCross    & \checkmark & 1.52          & \bf{3.02}          & 1.57          & 2.02          & \bf{29.76}                                    & 99.24          & 73.69 & \textbf{99.96}  & 99.99            \\
\midrule
Wang \et \cite{Wang_2022_CVPR}       & \checkmark & \checkmark & \checkmark & \checkmark & \checkmark & ---           & \textbf{2.78} & ---           & ---           & 111.65                                   & \textbf{99.61} & ---   & 99.35           & ---              \\
\bf{\ours}                                 & \checkmark & \checkmark & \checkmark & \checkmark & \checkmark & 1.16          & 2.58          & 1.23          & 1.77          & \textbf{29.84}                           & 99.33          & 72.70 & \textbf{99.89}  & 99.99            \\       
\bottomrule
\end{tabular}
}
\vspace{-0.2cm}
\caption{\textbf{Comparison on PROX} with Wang~\et~\cite{wang2020synthesizing} and Wang~\et~\cite{Wang_2022_CVPR}. Contact score threshold is 0.01. Results use first and last pose conditioning to match the compared SOTA. Results are refined by an optimization step.
\vspace*{-0.4cm}
} 

\label{tab:sota:prox2}
\end{table}

We compare our method to the state of the art~\cite{wang2020synthesizing,Wang_2022_CVPR} on PROX in Table~\ref{tab:sota:prox2}. Different sets of metrics and conditionings have been reported in the literature. To make a fair comparison, we match the conditioning with each method being compared.

We observe that our model provides a good trade-off between diversity (APD), quality (FD), and physical plausibility. In particular, non-collision scores for \ours does not vary substantially from the competing approaches while our model consistently has the highest contact score, which indicates a rich interaction with the scene. Simultaneously, a low FD score is achieved, which is a measure of realistic generations. 
To provide a more comprehensive understanding of how our method compares to these baselines, we have included qualitative video results in the supplementary material.

\subsection{Limitations}
\label{sec:limitations}
\vspace*{-0.1cm}
Since our method is purely kinematic, \ie, it does not take into account physics constraints, interpenetrations with the scene may occur in some cases. Furthermore, HUMANISE is a synthetic dataset and provides short motion clips that may not comprehensively capture all the subtleties associated with interacting with objects. 
We leave the use of more realistic datasets such as SAMP~\cite{hassan_samp_2021} for future work.

\section{Conclusion} 
\label{sec:conclusion}
\vspace{-1mm}
In this work, we introduce \ours,  a novel approach grounded in neural discrete representation for generating human motion within 3D virtual scenes.
\ours  can generate realistic motions while also capturing human-object interactions. 
This is an important step forward due to its potential applications in indoor activity simulation and synthetic data creation, among others.
Experiments show that \ours consistently improves upon the baselines. Additionally, our model can be controlled semantically, generalizes to a variety of new scenes, and generates long-term motions even if only very short sequences are present in the training data.

\vspace{2mm}
\noindent{\bf Acknowledgements}: \small{This work it partly supported by the  project MoHuCo PID2020-120049RB-I00 funded by MCIN/AEI/10.13039/501100011033. 
}

{
    \small
    \bibliographystyle{ieeenat_fullname}
    \bibliography{references}
}

\end{document}